\documentclass[twocolumn,11pt]{article}

\usepackage{graphicx}    
\usepackage{amsmath}     
\usepackage{geometry}    
\usepackage{hyperref}    
\usepackage{times}
\usepackage[switch]{lineno}      
\usepackage{subcaption}  
\usepackage{tcolorbox}
\tcbuselibrary{listings, breakable}
\usepackage[numbers]{natbib}

\begin{document}

\title{Modular Prompt Optimization: Optimizing Structured Prompts with Section-Local Textual Gradients}

\author{
Prith Sharma \quad Austin Z. Henley \\
Carnegie Mellon University \\
\texttt{priths@andrew.cmu.edu, azhenley@cmu.edu}
}

\maketitle


\section{Introduction}

Large language models (LLMs) are increasingly used as general-purpose problem solvers, where behavior is controlled primarily through prompting rather than task-specific training \cite{radford2019rewon,brown2020language}. In practice, prompt quality often becomes the main limiting factor. Small changes in wording, order, or formatting can lead to large differences in model performance, reliability, and even safety. This sensitivity is especially challenging for smaller, open-source instruction-tuned models, which typically require clearer instructions and benefit more from explicit structure \cite{belcak2025small}. As a result, developing principled and repeatable methods for improving prompts has become an important part of building robust LLM-based systems.

Real-world prompts are rarely simple or well-structured. They are often written as long blocks of text that mix several different purposes, such as defining the model’s role, providing background information, describing the task, listing constraints, and specifying the desired output format \cite{tian2025taxonomy}. When these elements are combined without clear separation, a range of problems can occur. Constraints may be incomplete or contradicted later in the prompt, formatting instructions may be ignored, and changes made to fix one issue can unintentionally introduce new ones. Over repeated rounds of refinement, prompts may also grow longer and harder to manage. These problems are difficult to diagnose because it is often unclear which part of the prompt is responsible for a given error, making prompt improvement slow and largely ad hoc.

Many prior works have studied automatic prompt optimization, but existing methods do not fully resolve these challenges. Continuous prompt tuning approaches optimize prompts in a latent embedding space while keeping the language model fixed \cite{shin2020autoprompt,lester2021power,li2021prefix}. Although this makes them parameter-efficient, the resulting prompts are not directly interpretable. More recent methods optimize prompts directly in natural language by treating improvement as a search or self-refinement process \cite{zhou2022large,pryzant2023automatic,yuksekgonul2024textgrad}. These approaches preserve readability and work in black-box settings, but they typically treat the prompt as a single piece of text and apply updates to it as a whole. As a result, it is hard to target specific parts of the prompt, interactions between unrelated instructions can increase, and important information may be altered during rewriting.

At the same time, practical prompting guidelines have increasingly adopted structured prompt formats that separate system role, context, task description, constraints, and output format \cite{openai_gpt5_prompting_2025,anthropic_claude_prompting_2025}. This practice reflects a simple but often overlooked idea that different parts of a prompt play different roles. When prompts are structured in this way, errors and improvements can be traced back to specific sections, and optimization can focus on refining individual components rather than rewriting the entire prompt at once.

In this paper, we introduce Modular Prompt Optimization (MPO), a schema-based prompt optimization framework that treats prompts as structured objects composed of fixed semantic sections. MPO applies section-local textual gradients at each iteration using a critic model which generates targeted natural-language improvements for each section independently while keeping the overall schema fixed. Updates are then consolidated with de-duplication to prevent prompt bloat and reduce interference between components. This design makes optimization localized, interpretable, and robust to uncontrolled rewriting.

We evaluate MPO on two reasoning benchmarks, ARC-Challenge \cite{allenai:arc} and MMLU \cite{hendryckstest2021,hendrycks2021ethics}, using LLaMA-3 8B-Instruct and Mistral-7B Instruct as solver models. Across both benchmarks and models, MPO consistently outperforms an untuned structured prompt and TextGrad \cite{yuksekgonul2024textgrad}. These improvements are achieved without modifying model parameters or altering prompt structure. The results suggest that maintaining a fixed prompt schema while optimizing semantic sections independently is an effective and practical approach for improving reasoning performance in small open-source LMs.

\section{Related Works}

\subsection{Continuous Prompt Optimization}
The first wave of automatic prompt optimization treated prompts as learnable parameters rather than linguistic objects. Methods such as \cite{shin2020autoprompt}, \cite{lester2021power}, \cite{li2021prefix}, and  \cite{liu2022p} optimized continuous prompt embeddings via gradient descent while freezing the underlying language model. More recent work continues to explore gradient-based optimization in embedding space \cite{hou2025embedgrad}.

While effective in parameter efficiency, these approaches produce prompts that are non-interpretable, task-specific, and difficult to audit or reuse. Because the optimized prompts exist in latent embedding space, they do not correspond to explicit natural-language instructions. Surveys of prompt tuning and prompt engineering highlight these limitations, noting that embedding-based methods optimize latent control signals rather than explicit instruction semantics \cite{li2025survey,chen2025unleashing}.
\subsection{Discrete Prompt Optimization in Language Space}
These limitations motivated a shift toward discrete prompt optimization, where prompts are optimized directly in natural language. Early work on prompt mining and prompt programming demonstrated that textual prompts could be systematically searched and evaluated \cite{jiang2020can,reynolds2021prompt}. Automatic Prompt Engineer showed that LLMs themselves can generate and rank candidate prompts, framing prompt optimization as a search problem in language space \cite{zhou2022large}.

Subsequent methods explored more structured optimization strategies, including beam-search-based updates and evolutionary algorithms over discrete prompts \cite{pryzant2023automatic,secheresse2025gaapo}. Modular frameworks such as \cite{zehle2025promptolution} further formalized prompt optimization as a compositional process over prompt components. These approaches preserve interpretability and enable black-box optimization but typically treat the prompt as a single monolithic block of text.
\subsection{Textual Gradients and Self-Refinement}
A parallel line of work reframed prompt optimization as an iterative self-improvement process, where language models critique and revise prompts using natural-language feedback. This includes self-refinement and reflection-based methods \cite{madaan2023self,shinn2023reflexion}, which were later formalized under the notion of textual gradients. In this paradigm, linguistic feedback plays a role analogous to gradient information in continuous optimization.

Representative methods include \cite{lin2024prompt}, \cite{pryzant2023automatic}, \cite{yuksekgonul2024textgrad}, \cite{cui2024introducing}, and reflective prompt evolution approaches such as \cite{agrawal2025gepa}. These techniques enable model-agnostic optimization while maintaining interpretability. However, feedback is generally applied globally to the entire prompt, making it difficult to attribute failures to specific semantic components or to ensure that critical information is preserved during optimization.

\subsection{Prompt Structure, Task Decomposition and positioning of our work}
Concurrently, practitioner-oriented prompting guides have converged on a consistent structured prompt format, typically decomposing prompts into semantic sections \cite{openai_gpt5_prompting_2025}, \cite{anthropic_claude_prompting_2025}. While these documents are not framed as optimization research, they implicitly encode a theory of prompt structure, suggesting that different components of a prompt serve distinct functional roles and should be reasoned about independently.

Recent research has begun to formalize this intuition. Task Facet Learning \cite{juneja2025task} proposes decomposing prompts into facets and learning new task-specific components from labeled errors. While effective, this approach modifies the prompt structure itself and focuses on discovering new task knowledge from downstream performance signals.

In contrast, our work is inspired by structured prompting practices rather than task-facet discovery. We assume a fixed, documentation-driven prompt schema and study how optimizing each semantic section independently using textual-gradient style feedback can improve performance. Rather than discovering new task facets or altering prompt topology, we focus on refining how existing task information is organized, expressed, and constrained within a stable prompt structure. This design is particularly motivated by small open-source LMs, which benefit disproportionately from explicit structure and modular instruction decomposition \cite{belcak2025small}.
\begin{figure*}[t]
    \centering
    \includegraphics[width=\textwidth]{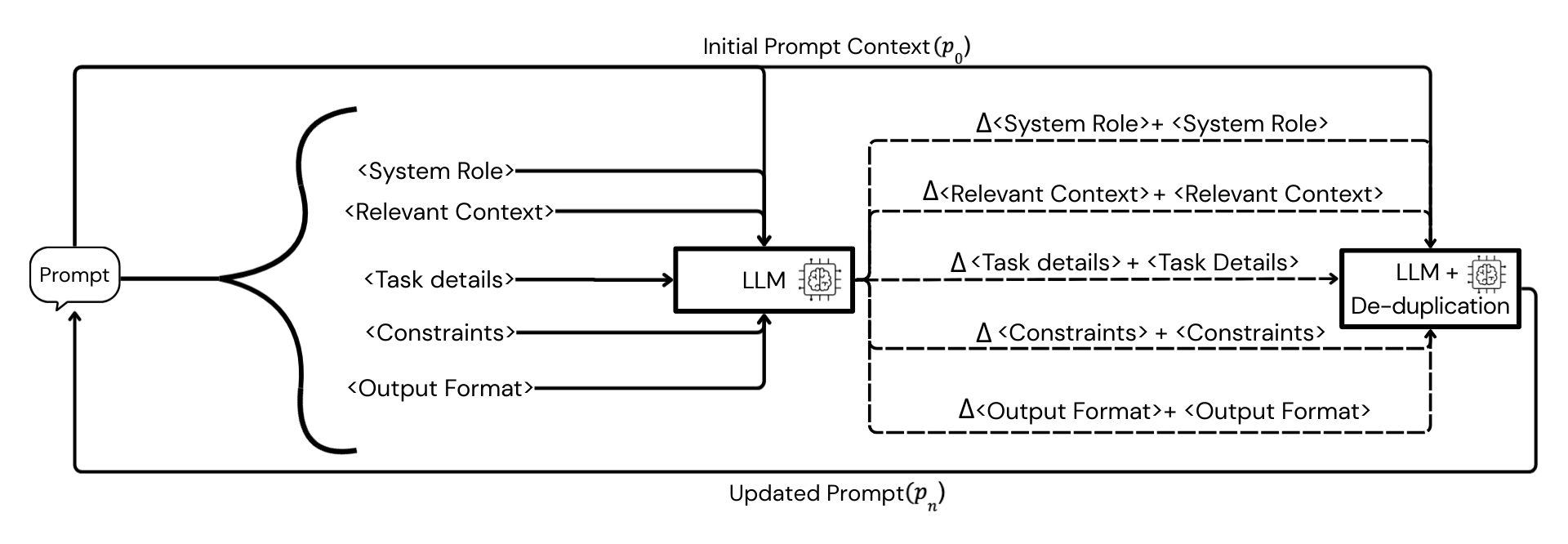}
    \caption{Overview of Modular Prompt Optimization (MPO). An initial prompt $p_0$ is decomposed into a fixed set of semantic sections, including system role, relevant context, task details, constraints, and output format. Each section is evaluated independently by a critic language model, which produces section-specific textual gradients. These gradients are combined with their corresponding original sections and passed through a de-duplication step to remove redundancy. The refined sections are then reassembled into an updated prompt $p_n$, preserving the original prompt structure while improving its content.}
    \label{fig:mpo_overview}
\end{figure*}
\section{Approach}
\label{sec:approach}

We propose Modular Prompt Optimization (MPO), a schema-based prompt optimization framework that treats prompts as structured objects composed of semantically distinct components. MPO refines prompts by iteratively improving the content of each component independently, while keeping the overall prompt schema fixed. An overview of the MPO optimization process is shown in Figure~\ref{fig:mpo_overview}.

\paragraph{Prompt schema.}
MPO decomposes each prompt into a fixed set of semantic sections: \emph{System Role}, \emph{Relevant Context}, \emph{Task Details}, \emph{Constraints}, and \emph{Output Format}. This decomposition is inspired by widely adopted practitioner prompting guidelines and reflects the observation that different parts of a prompt play distinct functional roles in shaping model behavior. Throughout optimization, the schema itself is held constant. MPO does not search over prompt structures or reorder sections. If a section is missing in the original prompt, it is initialized by extracting or inferring relevant content directly from the original prompt, without introducing new task information.

\paragraph{Prompt state and optimization objective.}
MPO maintains an explicit notion of prompt state. Let
\[
p_0 = \{ s_0^{(1)}, s_0^{(2)}, \dots, s_0^{(K)} \}
\]
denote the initial structured prompt, where each \( s_0^{(k)} \) corresponds to one semantic section. The goal of MPO is to produce a sequence of refined prompt states
\[
p_0 \rightarrow p_1 \rightarrow \dots \rightarrow p_n,
\]
where each successive prompt improves task performance while preserving interpretability and structural consistency.

\paragraph{Section-wise textual gradients.}
At each iteration \(t\), MPO evaluates each prompt section independently using a critic language model. Rather than rewriting the entire prompt, the critic produces textual gradients for each section. Formally, for section \(k\), the critic generates
\[
\Delta s_t^{(k)} = \mathcal{C}\!\left(s_t^{(k)},\, p_t \setminus s_t^{(k)}\right),
\]
where \(\mathcal{C}\) denotes the critic model and \(p_t \setminus s_t^{(k)}\) provides the surrounding prompt context without modifying other sections. Each \(\Delta s_t^{(k)}\) is expressed in natural language and represents a proposed improvement in alignment with the intended task.

\paragraph{Textual Gradients application and consolidation.}
Section updates are applied additively, yielding an intermediate section:
\[
\tilde{s}_{t+1}^{(k)} = s_t^{(k)} \oplus \Delta s_t^{(k)},
\]
where \(\oplus\) denotes textual aggregation. As repeated iterations may introduce redundancy or semantically overlapping instructions, MPO applies an LLM-based de-duplication and consolidation step:
\[
s_{t+1}^{(k)} = \mathcal{D}\!\left(\tilde{s}_{t+1}^{(k)}\right),
\]
where \(\mathcal{D}\) removes redundant content while preserving everything else. The updated prompt state is then reassembled as
\[
p_{t+1} = \{ s_{t+1}^{(1)}, s_{t+1}^{(2)}, \dots, s_{t+1}^{(K)} \}.
\]

\paragraph{Iterative refinement.}
This process can be repeated for a number of iterations depending on the prompt and task as shown in Figure~\ref{fig:mpo_overview}.. By restricting updates to localized, section-specific deltas and consolidating them after each iteration, MPO reduces interference between unrelated prompt components and prevents uncontrolled prompt growth. 

\paragraph{Experimental setup.}
We evaluate MPO on LLaMA-3 8B-Instruct and Mistral-7B-Instruct as solver models. For each model, we compare MPO-optimized prompts against Textgrad \cite{yuksekgonul2024textgrad} and a fixed baseline prompt that follows the same schema but does not undergo iterative refinement.

Evaluation is conducted on two standard reasoning benchmarks: ARC Challenge \cite{allenai:arc} and the full MMLU benchmark \cite{hendryckstest2021}, \cite{hendrycks2021ethics}. For ARC, performance is measured using exact-match accuracy on multiple-choice questions. For MMLU, we report average accuracy across all subject domains. 

MPO is applied independently for each model–dataset pair, producing a task-specific and model-specific optimized prompt. This setup allows us to assess both absolute performance gains and the consistency of MPO improvements across different models.

\begin{figure*}[t]
    \centering
    \includegraphics[width=\textwidth]{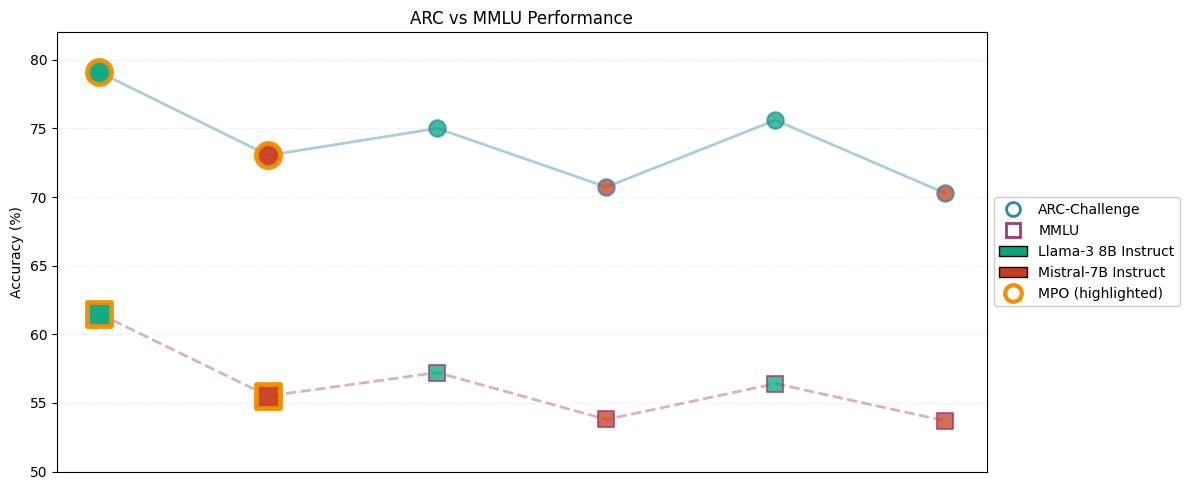}
    \caption{Accuracy on ARC-Challenge and MMLU for LLaMA-3 8B-Instruct and Mistral-7B-Instruct under different prompt optimization methods. Highlighted points denote MPO performance, which achieves higher accuracy than an untuned structured prompt and TextGrad across both benchmarks and models.}
    \label{fig:Results}
\end{figure*}

\section{Results}
\label{sec:results}

We evaluate Modular Prompt Optimization (MPO) on two reasoning benchmarks: ARC-Challenge and MMLU (all subjects). Experiments are conducted using two open-source instruction-tuned solver models: LLaMA-3 8B-Instruct and Mistral-7B-Instruct. We compare MPO against two baselines: (i) an untuned structured prompt that follows the same schema but is not optimized, and (ii) TextGrad

For ARC-Challenge, we use 1,119 training examples and 1,172 test examples. For MMLU, we use a training subset of 3,000 examples and evaluate on a held-out test set of 1,400 examples. 

\subsection{ARC-Challenge}

Table~\ref{tab:arc_results} reports ARC-Challenge test accuracy across models and optimization methods. 
Across both LLaMA-3 8B-Instruct and Mistral-7B-Instruct, MPO achieves the strongest ARC-Challenge performance, consistently surpassing both the untuned prompt and TextGrad.

\begin{table}[h]
\centering
\small
\begin{tabular}{lccc}
\hline
Model & Untuned & TextGrad & MPO \\
\hline
LLaMA-3 8B-Instruct & 75.0 & 75.6 & \textbf{79.10} \\
Mistral-7B-Instruct & 70.73 & 70.30 & \textbf{73.04} \\
\hline
\end{tabular}
\caption{ARC-Challenge test accuracy (\%) across solver models and prompt optimization methods.}
\label{tab:arc_results}
\end{table}

\subsection{MMLU (All Subjects)}

Table~\ref{tab:mmlu_results} presents results on MMLU across all subject domains. As with ARC-Challenge, MPO consistently improves performance across both solver models.
Notably, TextGrad degrades performance relative to the untuned prompt on MMLU. In contrast, MPO yields stable gains across models and tasks.

\begin{table}[h]
\centering
\small
\begin{tabular}{lccc}
\hline
Model & Untuned & TextGrad & MPO \\
\hline
LLaMA-3 8B-Instruct & 57.21 & 56.40 & \textbf{61.50} \\
Mistral-7B-Instruct & 53.79 & 53.70 & \textbf{55.50} \\
\hline
\end{tabular}
\caption{MMLU (all subjects) test accuracy (\%) across solver models and prompt optimization methods.}
\label{tab:mmlu_results}
\end{table}

Across both benchmarks and solver models, MPO consistently outperforms both the untuned prompt and TextGrad. These results suggest that enforcing a fixed prompt schema and applying section-local textual gradients leads to more stable and generalizable improvements than global updates.

\section{Conclusion and Future Work}
\label{sec:conclusion}

We presented Modular Prompt Optimization (MPO), a schema-constrained prompt optimization framework that refines prompts by independently optimizing their semantic components using section-local textual feedback followed by aggregation and de-duplication. By treating prompt structure as a fixed inductive bias rather than a free-form text object, MPO enables targeted refinement of system role, context, task specification, constraints, and output format without altering prompt topology or model parameters.

Across ARC-Challenge and MMLU, and over multiple open-source instruction-tuned models, MPO consistently outperforms both untuned structured prompts and global textual gradient-based methods such as TextGrad. These gains are achieved through localized updates that reduce interference between prompt components, leading to more stable and generalizable improvements. The results suggest that explicit prompt structure, coupled with section-wise optimization, is an effective mechanism for improving reasoning performance in open-source language models without additional training or task-specific data.

Future work can extend MPO along several directions. First, it would be interesting to study whether using stronger open-source models as critics for generating section-wise feedback leads to further improvements and how the choice of critic model affects optimization stability and transferability. Second, MPO offers a natural framework for studying prompt safety and robustness, including whether prompts that induce unsafe or jailbreak behaviors are amplified or mitigated through structured optimization, and how section-local updates interact with safety-aligned system roles and constraints. Finally, MPO may be evaluated on a broader range of reasoning benchmarks and coupled with automated validation or stopping mechanisms, and further combined with programmatic prompting or agentic pipelines to support more complex structured optimization workflows.


\bibliographystyle{abbrvnat}

\bibliography{references}     

\clearpage
\appendix

\onecolumn
\section{Prompt Optimization Illustration using Llama3 8B-Instruct}

\vspace{0.5em}

\noindent
\begin{minipage}[t]{0.48\textwidth}
\centering
\textbf{A \quad Sample Prompt Input}

\vspace{0.5em}

\begin{tcolorbox}[
  colback=white,
  colframe=black,
  boxrule=0.4pt,
  arc=2pt,
  left=4pt,right=4pt,top=4pt,bottom=4pt
]
\ttfamily\footnotesize
You will receive multiple intellectually unrelated questions below. You must answer them all, but you must do so while respecting the following meta-instructions, which may partially conflict.

Important instructions:
- Do not introduce information that is not logically required by the question.
- Do not provide unnecessary explanations unless explicitly requested.
- If a question can be answered at multiple levels of rigor, choose the minimal level that is still formally correct.
- Do not assume the reader is a beginner unless stated otherwise.
- Do not summarize across questions.

Rules to stick to:
- For mathematically defined objects, you may use standard notation without defining it.
- For physical concepts, provide interpretation but avoid speculative language.
- For computationally infeasible quantities, provide an exact representation if possible; otherwise, justify why an alternative representation is used.

Your tasks to perform would include:

1. State the general Taylor expansion of a function about a point a.
2. Explain the cosmological constant.
3. Determine the value of 100 factorial.
4. Compute 2 + 2.
5. Give me the complete table of 2, table of 5 and table of 10.
6. What is a transformer in NLP?

Would like output to be structured as follows:
- Answer all four tasks in a single response.
- Do not label the tasks or number your answers.
- Do not explicitly restate the questions.
- Do not include transitional phrases.
- Do not use bullet points or lists.
- For every task give me multiple questions for further thought.
\end{tcolorbox}
\end{minipage}
\hfill
\begin{minipage}[t]{0.48\textwidth}
\centering
\textbf{B \quad Structured Input Prompt}

\vspace{0.5em}

\begin{tcolorbox}[
  colback=white,
  colframe=black,
  boxrule=0.4pt,
  arc=2pt,
  left=4pt,right=4pt,top=4pt,bottom=4pt
]
\ttfamily\footnotesize
<System Role>
You are a helpful, creative, and smart assistant.

<Context>
You will receive multiple intellectually unrelated questions below. You must answer them all, but you must do so while respecting the following meta-instructions, which may partially conflict.

<Task>
1. State the general Taylor expansion of a function about a point a.
2. Explain the cosmological constant.
3. Determine the value of 100 factorial.
4. Compute 2 + 2.
5. Give me the complete table of 2, table of 5 and table of 10.
6. What is a transformer in NLP?

<Constraints>
- Do not introduce information that is not logically required by the question.
- Do not provide unnecessary explanations unless explicitly requested.
- If a question can be answered at multiple levels of rigor, choose the minimal level that is still formally correct.
- Do not assume the reader is a beginner unless stated otherwise.
- Do not summarize across questions.
- For mathematically defined objects, you may use standard notation without defining it.
- For physical concepts, provide interpretation but avoid speculative language.
- For computationally infeasible quantities, provide an exact representation if possible; otherwise, justify why an alternative representation is used.

<Output Format>
- Answer all four tasks in a single response.
- Do not label the tasks or number your answers.
- Do not explicitly restate the questions.
- Do not include transitional phrases.
- Do not use bullet points or lists.
- For every task give me multiple questions for further thought.
\end{tcolorbox}
\end{minipage}

\vspace{0.75em}

\vspace{1.2em}
\clearpage
\noindent
\textbf{C \quad Final Refined Prompt}

\vspace{0.5em}

\begin{tcolorbox}[
  colback=white,
  colframe=black,
  boxrule=0.4pt,
  arc=2pt,
  left=4pt,right=4pt,top=4pt,bottom=4pt
]
\ttfamily\footnotesize
You will receive multiple intellectually unrelated questions below. You must answer them all, but you must do so while respecting the meta-instructions and rules to stick to.

Important instructions: - Do not introduce information that is not logically required by the question. - Do not provide unnecessary explanations unless explicitly requested. - If a question can be answered at multiple levels of rigor, choose the minimal level that is still formally correct. - Do not assume the reader is a beginner unless stated otherwise. - Do not summarize across questions.

State the general Taylor expansion of a function about a point a. Explain the cosmological constant. Determine the value of 100 factorial. Compute 2 + 2. Give me the complete table of 2, table of 5 and table of 10. What is a transformer in NLP? Provide a single response without labeling or numbering, and include multiple questions for further thought for each task.

Answer all four tasks in a single response. Do not label the tasks or number your answers. Do not explicitly restate the questions. Do not include transitional phrases. Do not use bullet points or lists.
\end{tcolorbox}

\vspace{1.2em}

\noindent
\textbf{D \quad Model Responses}

\vspace{0.5em}

\textit{Response to initial prompt}
\begin{tcolorbox}[
  colback=gray!5,
  colframe=black,
  boxrule=0.4pt,
  arc=2pt,
  left=4pt,right=4pt,top=4pt,bottom=4pt
]
\ttfamily\footnotesize
The Taylor expansion of a function f(x) about a point a is given by:

$f(x) = f(a) + f'(a)(x-a) + (1/2)f''(a)(x-a)^2 + (1/6)f'''(a)(x-a)^3 +...$

The cosmological constant is a fundamental constant of the universe that represents the energy density of the vacuum. It is denoted by the Greek letter lambda ($\lambda$) and is a measure of the energy density of the vacuum. The cosmological constant is a key component of the Friedmann equations, which describe the evolution of the universe.

100! is equal to 100 factorial, which is the product of all positive integers from 1 to 100. This is a very large number, and its exact value is:

\[
100! = 9.33262154 \times 10^{157}
\]

2 + 2 is equal to 4.

The table of 2 is a multiplication table that shows the product of 2 multiplied by each integer from 1 to 10:

2 × 1 = 2
2 × 2 = 4
2 × 3 = 6
2 × 4 = 8
2 × 5 = 10
2 × 6 = 12
2 × 7 = 14
2 × 8 = 16
2 × 9 = 18
2 × 10 = 20

The table of 5 is a multiplication table that shows the product of 5 multiplied by each integer from 1 to 10:

5 × 1 = 5
5 × 2 = 10
5 × 3 = 15
5 × 4 = 20
5 × 5 = 25
5 × 6 = 30
5 × 7 = 35
5 × 8 = 40
5 × 9 = 45
5 × 10 = 50

The table of 10 is a multiplication table that shows the product of 10 multiplied by each integer from 1 to 10:

10 × 1 = 10
10 × 2 = 20
10 × 3 = 30
10 × 4 = 40
10 × 5 = 50
10 × 6 = 60
10 × 7 = 70
10 × 8 = 80
10 × 9 = 90
10 × 10 = 100

A transformer in NLP is a type of neural network architecture that is used for natural language processing tasks. It is a type of encoder-decoder model that uses self-attention mechanisms to process input sequences and generate output sequences. The transformer architecture was introduced in a 2017 paper by Vaswani et al. and has since become a widely used and highly effective architecture for many NLP tasks.

Further thought questions:

* What are some of the key advantages of the transformer architecture over other neural network architectures?
* How does the self-attention mechanism in the transformer work?
* What are some of the most common applications of the transformer architecture in NLP?

\end{tcolorbox}

\vspace{0.8em}
\clearpage
\textit{Response to refined prompt}

\begin{tcolorbox}[
  colback=gray!5,
  colframe=black,
  boxrule=0.4pt,
  arc=2pt,
  left=4pt,right=4pt,top=4pt,bottom=4pt
]
\ttfamily\footnotesize

The general Taylor expansion of a function $f(x)$ about a point a is:
\[
f(x) = f(a) + f'(a)(x-a) + (1/2)f''(a)(x-a)^2 +...
\]
The cosmological constant is a term in the equation of state of the universe that represents the energy density of the vacuum. It is often denoted by the Greek letter lambda ($\lambda$) and is a measure of the resistance to the expansion of the universe.

\[
100! = 100 × 99 × 98 ×... × 3 × 2 × 1 = 9.33262154 \times 10^{157}
\]

2 + 2 = 4

Here are the tables:

Table of 2:
2, 4, 6, 8, 10,...

Table of 5:
5, 10, 15, 20, 25,...

Table of 10:
10, 20, 30, 40, 50,...

In NLP, a transformer is a type of neural network architecture that is particularly well-suited for natural language processing tasks. It is based on self-attention mechanisms that allow the model to focus on different parts of the input sequence simultaneously.

Further thought questions:

* What are some common applications of the Taylor expansion in physics and engineering?
* How does the cosmological constant affect the behavior of the universe on large scales?
* What are some of the challenges in computing large factorials, and how are they typically addressed?
* How do transformers improve upon traditional recurrent neural networks in NLP tasks?
* What are some potential limitations of the transformer architecture, and how are they being addressed in current research?

\end{tcolorbox}

\twocolumn

\end{document}